\documentclass[conference]{IEEEtran}
\IEEEoverridecommandlockouts
\usepackage{cite}
\usepackage{amsmath,amssymb,amsfonts}
\usepackage{algorithmic}
\usepackage{graphicx}
\usepackage{textcomp}
\usepackage{xcolor}
\usepackage[a4paper, total={184mm,239mm}]{geometry}

\usepackage{amsmath}
\usepackage{siunitx}
\usepackage[export]{adjustbox}

\def\BibTeX{{\rm B\kern-.05em{\sc i\kern-.025em b}\kern-.08em
    T\kern-.1667em\lower.7ex\hbox{E}\kern-.125emX}}
\begin{document}

\title{
Land \& Localize: An Infrastructure-free and Scalable Nano-Drones Swarm with UWB-based Localization
}
\author{
\IEEEauthorblockN{
Mahyar Pourjabar\IEEEauthorrefmark{1}, 
Ahmed AlKatheeri\IEEEauthorrefmark{2},
Manuele Rusci\IEEEauthorrefmark{3},
Agata Barcis\IEEEauthorrefmark{2}
}
\IEEEauthorblockN{
Vlad Niculescu\IEEEauthorrefmark{4},
Eliseo Ferrante\IEEEauthorrefmark{2},
Daniele Palossi\IEEEauthorrefmark{5}\IEEEauthorrefmark{4}
Luca Benini\IEEEauthorrefmark{1}\IEEEauthorrefmark{4}
}

\IEEEauthorblockA{\IEEEauthorrefmark{1} Department of Electrical, Electronic and Information Engineering (DEI), University of Bologna, Italy}
\IEEEauthorblockA{\IEEEauthorrefmark{2} Autonomous Robotics Research Center (ARRC), Technology Innovation Institute, UAE}
\IEEEauthorblockA{\IEEEauthorrefmark{3} Department of Electrical Engineering, KU Leuven, Belgium}
\IEEEauthorblockA{\IEEEauthorrefmark{4} Integrated Systems Laboratory (IIS), ETH Z\"urich, Switzerland}
\IEEEauthorblockA{\IEEEauthorrefmark{5} Dalle Molle Institute for Artificial Intelligence (IDSIA), USI-SUPSI, Switzerland}
Contact authors: mahyar.pourjabar2@unibo.it,
ahmed.alkatheeri@tii.ae
}

\maketitle

\begin{abstract}
Relative localization is a crucial functional block of any robotic swarm.
We address it in a fleet of nano-drones characterized by a \SI{10}{\centi\meter}-scale form factor, which makes them highly versatile but also strictly limited in their onboard power envelope.
State-of-the-Art solutions leverage Ultra-WideBand (UWB) technology, allowing distance range measurements between peer nano-drones and a stationary infrastructure of multiple UWB anchors.
Therefore, we propose an UWB-based \textit{infrastructure-free} nano-drones swarm, where part of the fleet acts as dynamic anchors, i.e., \textit{anchor-drones} (ADs), capable of automatic deployment and landing.
By varying the ADs' position constraint, we develop three alternative solutions with different trade-offs between flexibility and localization accuracy.
In-field results, with four flying \textit{mission-drones} (MDs), show a localization root mean square error (RMSE) spanning from \SI{15.3}{\centi\meter} to \SI{27.8}{\centi\meter}, at most.
Scaling the number of MDs from 4 to 8, the RMSE marginally increases, i.e., less than \SI{10}{\centi\meter} at most. The power consumption of the MDs' UWB module amounts to \SI{342}{\milli\watt}.
Ultimately, compared to a fixed-infrastructure commercial solution, our infrastructure-free system can be deployed anywhere and rapidly by taking \SI{5.7}{\second} to self-localize 4 ADs with a localization RMSE of up to 12.3\% in the most challenging case with 8 MDs.
\end{abstract}

\begin{IEEEkeywords} 
UAV, nano-drone swarm, ultra-wideband, onboard localization, indoor tracking 
\end{IEEEkeywords}

\section{Introduction} \label{sec:introduction}

A swarm of drones is made by multiple agents that can operate in a coordinated way to accomplish tasks efficiently, e.g., search and rescue missions and inspection~\cite{Waharte2009CoordinatedSW}.
The design of these robotic platforms becomes particularly challenging when targeting a swarm of fully-autonomous nano-drones.
With a form factor of \SI{10}{\centi\meter} and a limited on-board energy availability~\cite{DATE21_nano,Palossi2017TargetFO}, these versatile robots are severely constraint in sensing, processing, and communication resources, impacting the capabilities of each agent and the entire swarm. 

Accurate localization is at the functional core of any swarm operation, where each nano-drone must know its relative position to coordinate with the rest of the fleet.
An upcoming technology for precise localization on nano-drones is the Ultra-WideBand (UWB) distance ranging~\cite{niculescu2022energy, Queralta2020UWBbasedSF}.    
Every swarm's agent is equipped with a UWB module enabling relative distance estimations to other UWB devices using time-of-flight measurements~\cite{7303496}.
Traditional solutions based on UWB ranging can achieve high accuracy (i.e., \SI{10}{\centi\meter} error in ideal condition~\cite{Mazhar2017PreciseIP}) but rely on a \textit{fixed-infrastructure} of static nodes (i.e., anchors). 
Accordingly, their usability restricts to environments preliminarily equipped with ad-hoc UWB modules. 
Furthermore, a ranging accuracy degradation of up to \SI{35}{\centi\meter} has been reported when flying agents move outside the delimited space by the anchor nodes~\cite{Grasso2022AnalysisAA} 

\begin{figure}[t]
    \includegraphics[width=\columnwidth]{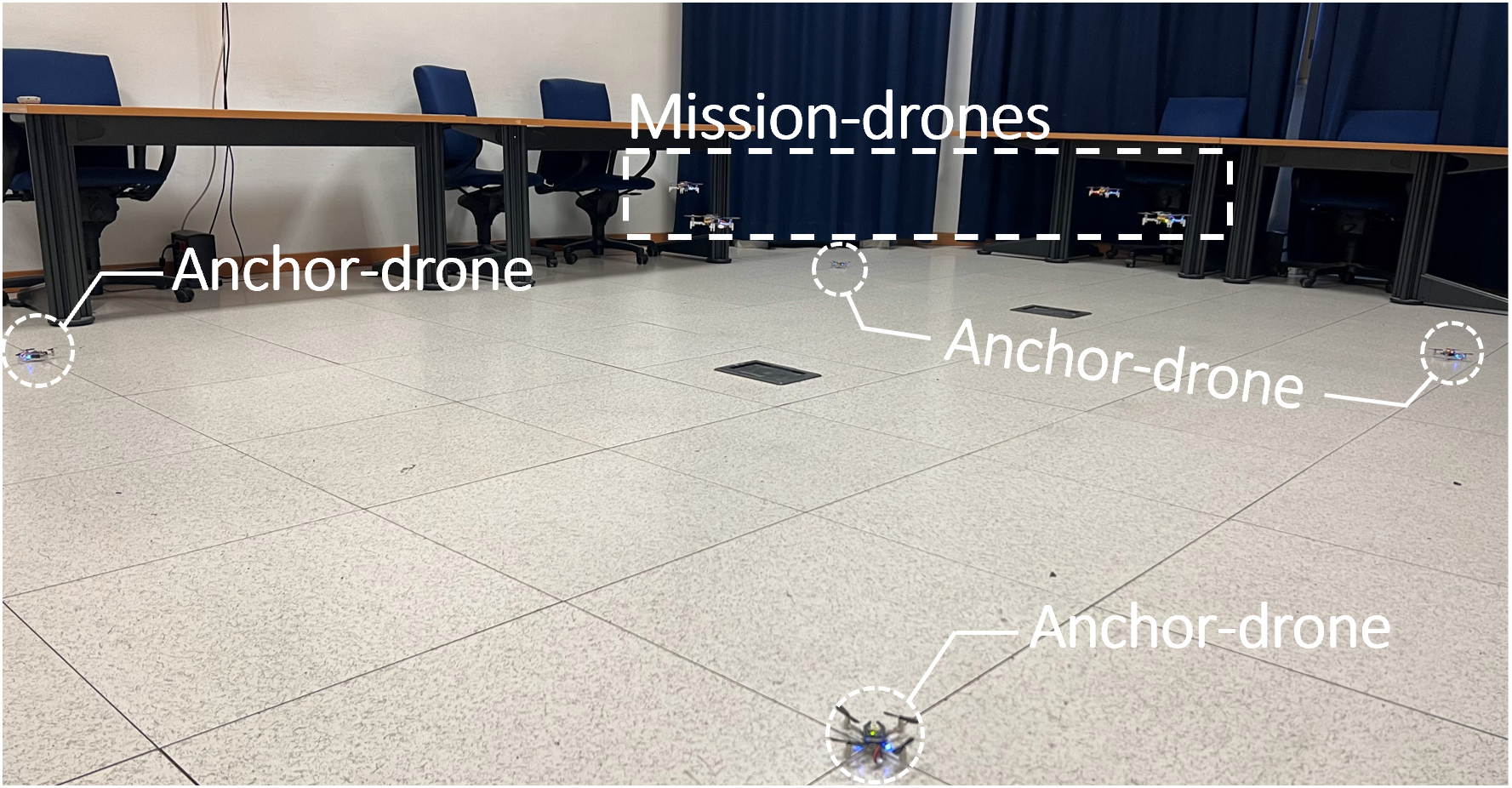}
    \caption{In-field deployment of our \textit{infrastructure-free} swarm, with highlighted 4 anchor-drones and 4 mission-drones.}
    \label{fig:proto}
\end{figure}

In this work, we tackle the limitation of static infrastructures by investigating alternatives based on the fundamental idea that a few nano-drones of the swarm dynamically land to act as UWB anchors for the rest of the fleet. 
We use this approach to eliminate the need for fixed module installation in the operation space, i.e., no anchor node needs to be pre-installed in the mission environment.
Therefore, we refer to our solution as \textit{infrastructure-free} localization.
In this setup, we define \textit{Anchor Drones} (ADs) as those who serve the localization of the other agents, which are called \textit{Mission Drones} (MDs), as shown in Fig.~\ref{fig:proto}.

Our contributions are in the design and in-field comparison of three systems based on different constraints on the knowledge of the ADs landing location.
In the \textit{ideal-landing} case, we assume the ADs to land precisely in their expected locations.
Then, in the \textit{automatic-landing} case, only the take-off position of the ADs is precisely known, while the landing spot suffers from the unavoidable onboard state estimation error during a $\sim$\SI{2}{\meter} autonomous flight.
Finally, in our third scenario, called \textit{self-localizing}, neither the take-off point nor the landing one is a-priori known; therefore, after the take-off, the ADs run aboard a self-localization routine to estimate the current position before navigating to their landing spot.
To enable these alternative solutions, we also provide a novel software library that contains the primitives to efficiently realize UWB-based communication, ranging, and synchronization between ADs and MDs. 

Our in-field evaluation employs 4 ADs, while we show the scalability of our localization systems by varying the number of MDs from 4 to 8. 
Employing 4 MDs, the localization Root Mean Square Error (RMSE) of our systems tasked with a predefined circular trajectory vary from \SI{15.3}{\centi\meter} up to \SI{27.8}{\centi\meter} for the ideal landing and self-localizing, respectively.
The localization RMSE increases by 31\% on average when scaling the number of MDs to 8 and affects the in-field control RMSE, i.e., the error between the desired and the actual trajectory, by \SI{13.1}{\centi\meter}, at most.
Compared with a commercial off-the-shelf (COTS) system using a UWB static infrastructure (6 anchors), our flexible infrastructure-free designs show a limited localization degradation by adding \SI{26.1}{\centi\meter} to the localization RMSE, at most, in the case of 8 MDs.
Finally, we provide onboard execution insights, such as \SI{5.7}{\second} to run a full round of ADs' self-localization routine and \SI{342}{\milli\watt} UWB power consumption for each agent in a swarm of 8 MD.
\section{Related Work} \label{sec:related_work}

The problem of relative localization in robotics has been addressed by employing various onboard sensors, e.g., GPS, LIDAR, cameras, etc.~\cite{Chen2022ASO}.
However, many of them are unsuitable for nano-drones due to their power envelope and physical constraints (i.e., payload and form factor)~\cite{DATE17_nano}.
Among the approaches that meet form factor and payload constraints, GPS is not usable as most nano drone deployment scenarios are indoor.
Nano-drones estimating their ego-motion usually fuse Inertial Measurement Units (IMU) information with visual cues (e.g., based on optical flow)~\cite{Palossi2017TargetFO}.
Even accepting their unavoidable unbounded drift~\cite{Balamurugan2016SurveyOU}, these methods would be inadequate for a swarm that requires a common reference frame.
Similarly, pure vision-based solutions for nano-drones~\cite{Li2022SelfsupervisedMM} are particularly challenging due to their computational requirements and all the limitations of vision~\cite{Lu2018ASO}, such as motion blur, occlusions, saturation, and exposure.

Therefore, many recent studies investigate UWB-based localization~\cite{Queralta2020UWBbasedSF,zhao2021learning}, primarily focusing on static infrastructure that exploits multiple fixed UWB anchors to localize swarm's agents within their receptive range~\cite{almansa2020autocalibration,hyun2019uwb,hamer2018self}. 
In~\cite{SnapLoc}, the authors achieve a localization error of \SI{33}{\centi\meter} by pushing the UWB ranging rate toward the theoretical upper bound of \SI{2.3}{\kilo\hertz} thanks to concurrent ranging from multiple devices, but without reporting for the UWB power characterization.
\cite{KhalafAllah2020ParticleFF} proposes a particle filtering algorithm for UWB localization based on Time Difference of Arrival (TDoA).
This work, tested in simulation, achieves a localization performance of \SI{11}{\centi\meter} and \SI{25}{\centi\meter} in a 2D and 3D scenario, respectively.
Zhao et al. introduce a bias correction and M-estimation-based outliers filter on nano-drones capable of reducing the UWB ranging noise by 42\%~\cite{zhao2021learning}.
However, all the methodologies above have the crucial limitation of relying on a pre-deployed infrastructure of UWB nodes positioned at known coordinates. 

Self-positioning systems have been proposed to mitigate this need for apriori knowledge of the anchors' position. 
Still, they require significant computational power, translating into a long execution time for constrained platforms, such as UWB anchor nodes.
Almansa et al.~\cite{almansa2020autocalibration}  present a self-localization procedure between 4 anchors based on a Decawave's DWM1001 Development board (i.e., ARM Cortex-M4 at \SI{64}{\mega\hertz}).
The basic system requires up to \SI{40}{\second} and scores a localization error above \SI{1}{\meter} when the anchors are placed less than \SI{20}{\meter} from each other. 
The authors further reduce the latency by proposing a custom auto-calibration algorithm based on a repetitive ranging mechanism, which reduces the latency by 40$\times$ and lowers the error by  17\%.
However, the mentioned auto-calibration system has been tested only in simulation and still addresses a static infrastructure setup.

Among the works investigating infrastructure-free solutions, G{\"u}ler et al. propose a UWB-based localization method featuring three UWB modules on a single land robot to localize the target robot~\cite{guler2018real}.
The same principle has been later applied to a hexacopter~\cite{guler2019infrastructure}.
Unfortunately, placing three beacons at a sufficient distance on nano-drones is not feasible.
On the contrary, Li et al. presented an approach for estimating the relative localization of a swarm of nano-drones by introducing a relative Kalman filter fed with the inter-drones range measurements~\cite{chiavicaDegliInferi}.
The proposed relative Kalman estimator converges within \SI{15}{\second} among three drones. 
However, this method suffers from a quadratic scaling cost for the UWB-ranging operations with respect to the number of mission drones, while our approach can scale linearly.

Mor{\'o}n et al.~\cite{moron2022towards} propose a dynamic switching policy between TDoA and Two Way Ranging (TWR) ranging methods to gain scalability in multi-drone localization.
For a single drone, they achieved a maximum trajectory tracking error of less than \SI{40}{\centi\meter}, which is higher than our measures.
Moreover, this work was developed and tested on an Intel x5-Z8350 quad-core processor at \SI{1.44}{\giga\hertz}.
In contrast, we design and compare three infrastructure-free localization systems for a swarm of nano-drones based on our lightweight UWB library, in which self-localization takes \SI{5.7}{\second} on a Cortex M4 single-core microcontroller unit (MCU).
\section{UWB-based localization} \label{sec:uwb_localization}

\subsection{Background} \label{sec:uwb_technology}

UWB is a novel technology that enables both data transfer and distance measurements (i.e., ranging) with an accuracy of \SI{10}{\centi\meter} in ideal conditions~\cite{Mazhar2017PreciseIP}.
Among the existing ranging schemes used with UWB, TWR is one of the most popular due to its high accuracy compared to the alternatives. 
In a TWR scheme between two devices, the first one, i.e., the \textit{initiator}, sends a message to the \textit{responder}, which replies with an acknowledge message.
Since both messages embed the timestamp information, it is possible to compute the round-trip time and, therefore the distance between the two modules, knowing that the wave travels with the speed of light.
The drawback of this scheme, called Single-Sided TWR (SS-TWR), is the high sensitivity to clock drifts.
The Double-Sided TWR (DS-TWR) scheme mitigates this issue by introducing a  third message sent from the initiator to the responder to compute the time of flight.
In DS-TWR, the responder calculates the range, and therefore a fourth data message exchange may be necessary to send back the range measurement to the initiator.

When empowered by UWB radios, nano-drones can localize themselves by ranging with respect to a set of UWB anchor-nodes placed at known positions.
On present solutions~\cite{Chen2022ASO}, each nano-agent uses an Extended Kalman Filter (EKF) to enable state estimation for our nonlinear dynamic system. 
EKF turns the sensor signals into an estimate of the current state over time, e.g., the current position.
Our system utilizes the \textit{Bitcraze EKF implementation} as a part of \textit{crazyflie-firmware}\footnote{https://github.com/bitcraze/crazyflie-firmware}.
The filter is fed with the ranging measurements, the coordinates of the UWB fixed nodes, and other sensor data, e.g., optical flow. 
The filter updates whenever a new ranging measure is available and returns the new estimated location. 

At present, the application software of these swarm localization solutions implements TWR operations by reading/writing the internal registers of the UWB radio chip (\textit{Decawave}'s UWB chips are nowadays a common choice). 
This severely affects the code readability and the portability of individual UWB functionalities to other application tasks. 
These limitations motivated us to propose a new modular UWB software library that includes multiple UWB primitives for a wide variety of UWB-based tasks for a multi-agent system. 

\begin{figure}[t]
    \centering
    \includegraphics[width=0.5\textwidth]{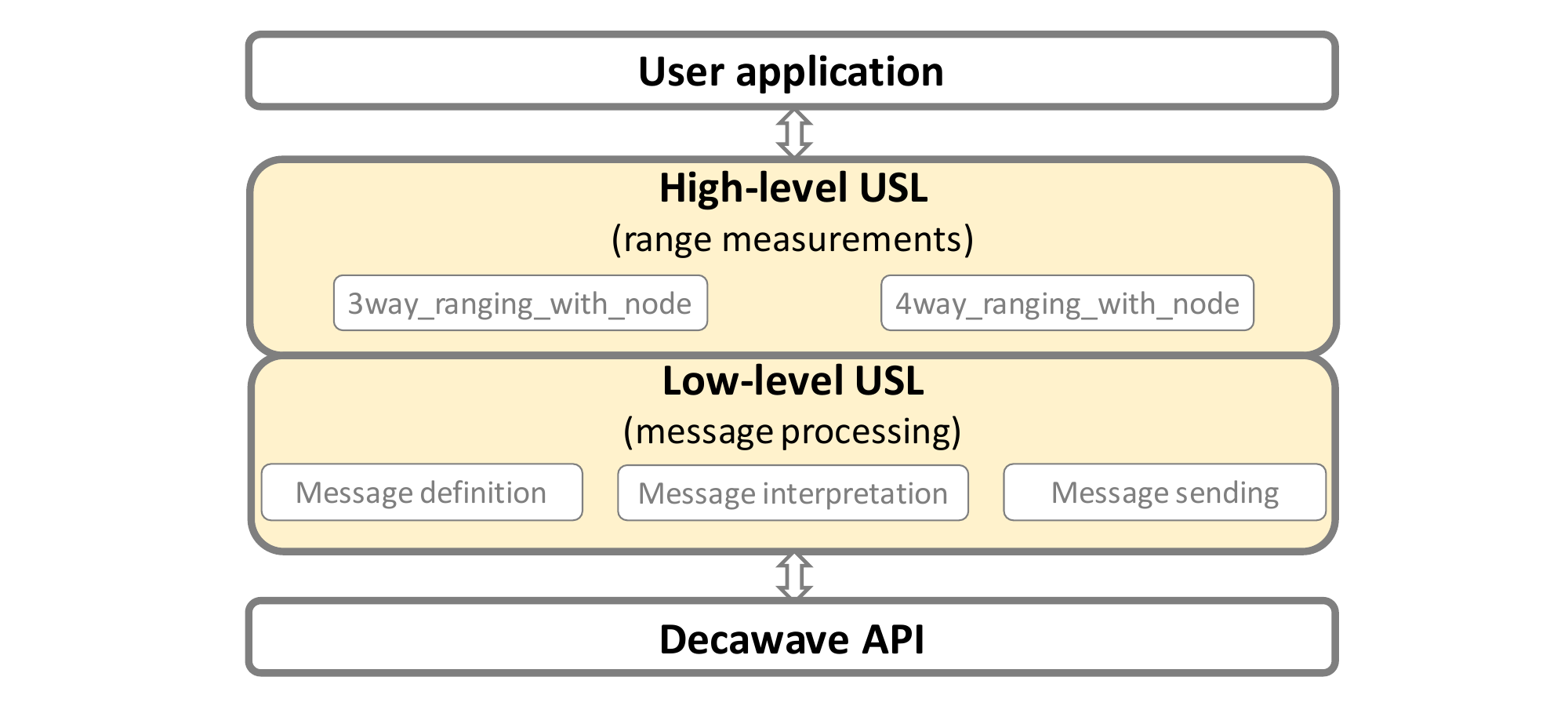}
    \caption{Components  of the UWB Software Library stack (in yellow).}
    \label{fig:UWB_API_sw_stack}
\end{figure}

\subsection{UWB Software Library} \label{sec:sw_lib}

The proposed open-source UWB Software Library (USL)\footnote{https://github.com/vladniculescu/uwb-software-library} manages the programming of the UWB radio chip in a user-transparent way to facilitate the development process. 
High-level functionalities are exposed to the programmer with simple function calls, including ranging and data communication between two agents.
As shown in Fig.~\ref{fig:UWB_API_sw_stack}, these functions are gathered into the High-level USL component and can be directly used within the application code. 
The basic routines at the core of the high-level functions, e.g., UWB message creation and interpreter, are collected into the Low-Level USL layer. 
These low-level functions make use of vendor-specific APIs, e.g., Decawave's register read/write operation, to efficiently exploit the underlying hardware.
As the main benefit, this modular library enables the rapid development of new functions without dealing with the complexity of the hardware settings, which are abstracted by the low-level USL routines.
In a multi-agent system with UWB modules, a unique ID is assigned to every device and can be accessed by the USL primitives.  
For exchanging data between two UWB nodes, the low-level USL defines the structure of a UWB message as follows: a control byte, the source ID, the destination ID, the message type, the data payload, and its length.
The value of the control byte is fixed (i.e., 0xDE) among all the entities running the USL, and it is used to filter out foreign messages. 
The source ID and destination ID refer to the sender and the receiver of the current message, respectively.
If a drone receives a message whose destination ID does not match its own ID, the message is ignored. 
The message type tells the USL how to interpret the message according to its purpose, e.g., an initial ranging message, a response message, etc. 
Finally, the data field embeds the payload associated with each type of message –- in the ranging case, this is normally filled with the timestamp values used to calculate the distance.

When sending data to another node, the low-level function \texttt{send\_msg} calls device-specific instructions to set the UWB radio chip in TX mode. 
After sending the data, the device can wait for a response (\texttt{send\_msg\_wait\_rsp}), which is useful when an acknowledgment is desired, or it can be simply put in listening mode (\texttt{wait\_msg}). 
In both cases, the radio RX mode is active. 
Typically, the UWB subsystem presents a higher average power consumption in RX mode than in the TX mode~\cite{Mayer2019EmbedUWBLP}. 
When transmitting, the radio amplifier circuit is only activated during the message-sending slots, whereas in RX mode, these circuits have to be constantly powered to receive the transmitted signals.

The High-level USL is implemented on top of the Low-level USL and includes simple function calls for ranging with another UWB node. 
For instance, the \texttt{3way\_ranging\_with\_node(srcID,
dstID)} function implements the standard DS-TWR by exchanging three UWB messages among the \texttt{srcID} and the \texttt{dstID} nodes. 
The \texttt{4way\_ranging\_with\_node} follows a similar  scheme, but at the end, the responder sends an additional UWB message to the initiator node to communicate the measurement back.

Lastly, the USL features an internal state variable that dictates if the UWB node operates as an initiator or if it is found in listening RX mode, 
where it operates as a responder. 
This state can change dynamically.
Normally, all devices of the fleet work as responders, and only one node switches the role from the responder to the initiator to start a ranging or communication operation. 
\section{Infrastructure-free Localization Design} \label{sec:sys_desing}

\begin{figure*}[t]
    \centering
    \includegraphics[width=1.9\columnwidth]
    {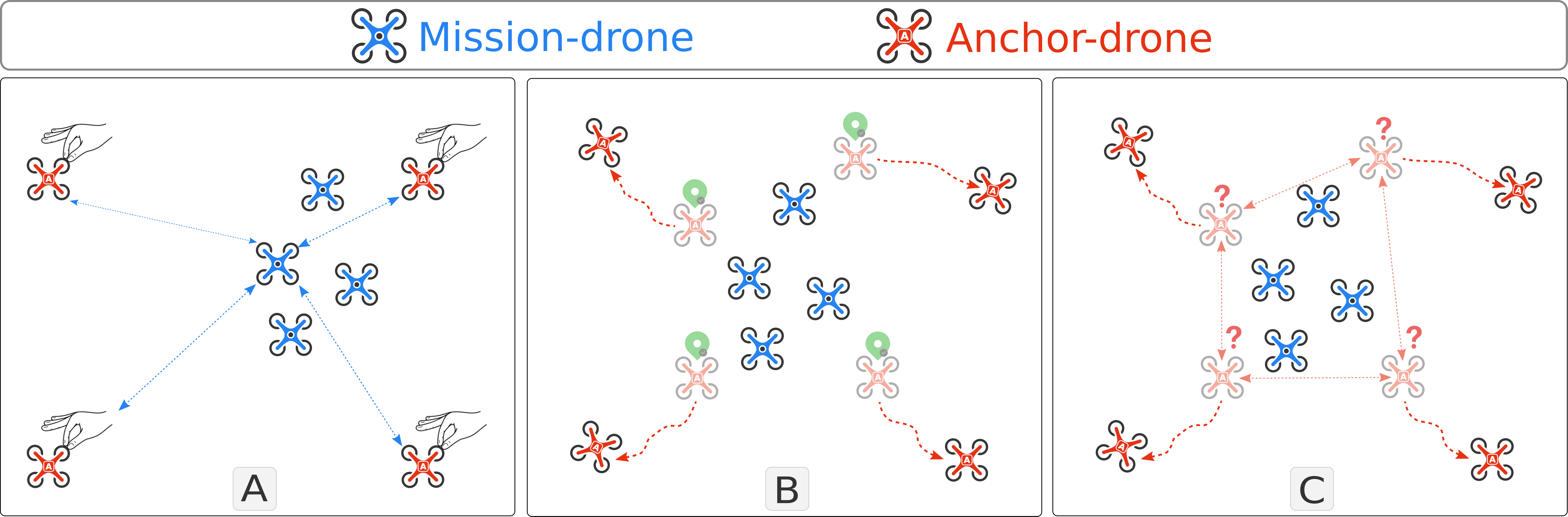}
    \caption{Proposed infrastructure-free localization systems. From left: ideal-landing, automatic-landing and self-localize.}
    \label{fig:dynamic_anchor_drone_placement}
\end{figure*}

The proposed localization systems rely solely on  a homogeneous swarm of nano-drones.
Every agent accommodates a UWB radio module, a low-power MCU, for control and onboard processing tasks, and a set of sensors: IMU, an Optical Flow sensor, and an altitude range sensor.
The MCU handles UWB ranging and communication tasks using the USL.

In our scheme, every drone can be configured as an Anchor Drone (AD) or a Mission Drone (MD). 
In general, we consider a swarm composed of $M$ ADs and $N$ MDs, and we refer to a specific agent as AD$i$, $i \in [0,1,\dotsc M-1]$, or MD$i$, $i \in [0,1,\dotsc N-1]$, respectively.
ADs and MDs differ in function. 
The former fly ahead autonomously to reach the assigned target landing positions $p^T_{ADi}$, which are expressed with respect to a global reference frame. 
During the flight, the ADs estimate the current position by feeding a Kalman Filter with the IMU and Optical Flow sensors data only. 
After landing, every nano-drone stores the estimated landing position $\tilde{p}^T_{ADi}$, which will be later included in the payload of the UWB messages. 
The actual landing position of nano-drones is denoted as $\tilde{p}^A_{ADi}$.

Once all ADs are landed, the MDs can localize themselves by ranging with the ADs, exploiting the same logic described in Sec.~\ref{sec:uwb_technology}. 
Hence, the ADs form a self-adaptive UWB infrastructure for the localization of the MDs. 
All the drones rely on the DS-TWR scheme and make use of the USL primitives for ranging.
Initially, all MDs and ADs are in listening mode  and behave as \textit{responders}.
The ranging process is triggered by AD$0$, which switches its internal state to \textit{initiator} and starts ranging with MD$0$.
The other ADs also receive this message and schedule the ranging with MD$0$ after a time delay that is a function of their index $i = 1,.., M-1$. 
More in detail, given a fixed ranging latency of \SI{3}{\milli\second}, AD$i$ switches on the UWB radio to perform ranging with MD0 after $i\times$ \SI{3}{\milli\second} from the beginning of the process, and then the radio is turned off again.
In our implementation, every AD knows the number of MDs involved in the ranging.
Note that, this parameter can be changed dynamically, e.g., by means of radio broadcast communication.

In the following, we describe three localization system designs that distinguish from the different estimations of $\tilde{p}^T_{ADi}$.

\subsection{Ideal-Landing} \label{sec:ideal_landing}
The first scenario considers $\tilde{p}^T_{ADi} = p^T_{ADi} = p^A_{ADi}$, meaning the ADs communicate the \textit{precise} landing positions to the MDs for the localization task.
As shown in Fig.~\ref{fig:dynamic_anchor_drone_placement}A, this setup is realized by manually placing the ADs on the target land coordinates, mainly used for comparisons with the other systems.

\subsection{Automatic-Landing} \label{sec:automatic_landing}
The ADs fly to their target landing positions after taking off from known locations, as represented in Fig.~\ref{fig:dynamic_anchor_drone_placement}B. 
To this end, we initialize the ADs' state estimators with the actual initial positions. 
The internal Kalman filter computes the current location by taking only the IMU and Optical Flow sensor data as inputs. 
The drone's flight controller uses this information to navigate to $p^T_{ADi}$. 
After landing, $\tilde{p}^T_{ADi}$ is updated with the last output of the Kalman Filter. 
The landing error $| p^A_{ADi} - \tilde{p}^T_{ADi} |$ is measured as the distance between the actual landing position and the internally estimated landing point.
This error is affected by two primary sources: (i) the sensors' noisy measurements (e.g., Optical Flow can be highly noisy) and (ii) imperfect take-off and landing procedures, e.g., drones bouncing on the ground during landing. 

\begin{figure}[b]
    \centering
    \includegraphics[width=1\columnwidth]{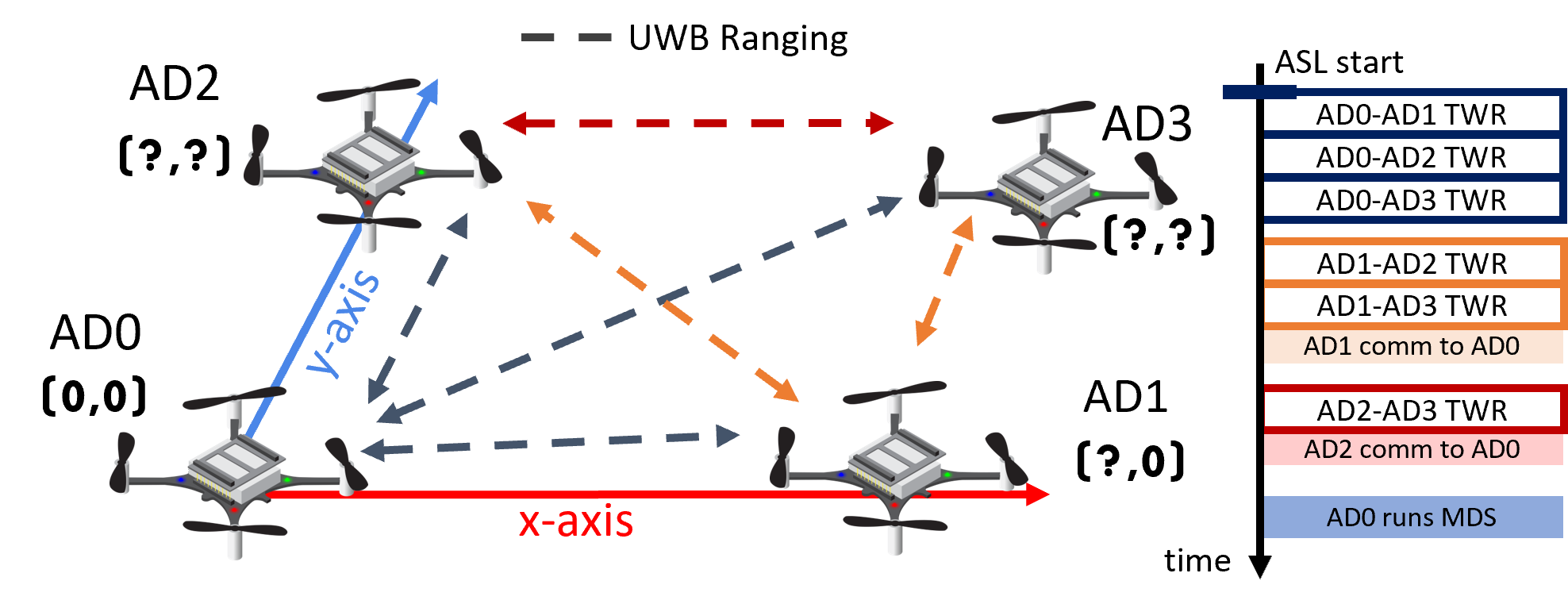} 
    \caption{Anchor Self-Localization (ASL) process. The ADs perform UWB ranging according to the time schedule shown on the right. AD0 collects all the measures before running the MDS algorithm.}
    \label{fig:self-localization-flow}
    \vspace{-1em}
\end{figure}

\subsection{Self-Localizing} \label{sec:self_localization}

To eliminate the previous constraints of knowing the ADs' precise landing coordinates (Ideal-Landing) and of the known initial take-off position (Automatic-Landing), we boost the previous system with an \textit{Anchors Self-Localization} (ASL) procedure to estimate the relative position of the ADs based only on the distance between them.

Initially, every AD tracks its location with respect to a local reference frame. To start the ASL task, all ADs reduce their velocity to \SI{0}{\meter/\second} and keep hovering.
Within our architecture with $M$ ADs, AD$0$ drives the process. 
The ASL works in two steps. 
First, a distance matrix $D$ is obtained by measuring the distances between the ADs via UWB ranging.
Then, AD0 uses the  Multi-Dimensional Scaling (MDS) algorithm \cite{MDS_algo} to derive the ADs' relative localization based on the distance matrix.

As shown in Fig.~\ref{fig:self-localization-flow}, the relative coordinate system is fully defined by three anchor drones:
AD$0$ defines the origin, AD$0$ - AD$1$ is the X axis, and AD$0$ - AD$2$ is the direction of the Y axis.
In other words, the coordinates of the anchors AD$0$, AD$1$, and AD$2$ are (0,0), ($x\sb{1}>0$, 0), and ($x\sb{2}$, $y\sb{2}>0$), respectively.
These positions will be determined by the ASL together with the coordinates of other anchors ADi, i=3,...M-1.

The distance matrix acquisition scheme is designed so that AD$i$ does ranging with all the anchors with higher index (i.e., $i+1,i+2,\dotsc M-1$). 
Initially, AD$0$ switches its UWB state to the initiator to measure the distance with respect to other ADs, which work as responders. 
After a ranging exchange, every anchor AD$i$, $i>0$, schedules a ranging round after a period of $i \cdot \frac{(2 \cdot M - i - 1)}{2} \cdot \Delta t \cdot n\sb{meas}$.
Here, $\frac{(2 \cdot M - i - 1)}{2}$ represents the number of anchors that AD$i$ has to do ranging with, $\Delta t = \SI{4}{\milli \second}$ represents the duration to acquire a single range measurement at both sides of the ranging parties, and $n\sb{meas}$ represents the total number of measurements acquired between any two ADs. 
Due to the noisy nature of the UWB measurements, measuring the distance between two anchors is done by averaging over $n\sb{meas}=100$ UWB measurements.
When an AD finishes ranging with the other ADs, it sends the acquired (averaged) measurements to AD0 to be added to the distance matrix.

AD$0$ feeds the distance matrix into the MDS algorithm, which retrieves the anchors' coordinates using the singular value decomposition method.
More in detail, the coordinate matrix is computed as the product between the matrix of eigenvectors and the square root of the diagonal eigenvalue matrix.
The obtained coordinates are rotated so that AD$0$ and AD$1$ are both on the X-axis and translated so that AD$0$ is in origin.
In order to perform the eigenvalue decomposition, we use \textit{Eigen}, a lightweight C++ library that we link to our application.
Once the computation is done, AD0 shares the computed coordinates with the other ADs, and then all fly toward the predefined target landing points.
\section{Experimental Results} \label{sec:results}

\begin{figure}[t]
    \centering
    \includegraphics[width=1\columnwidth]
    {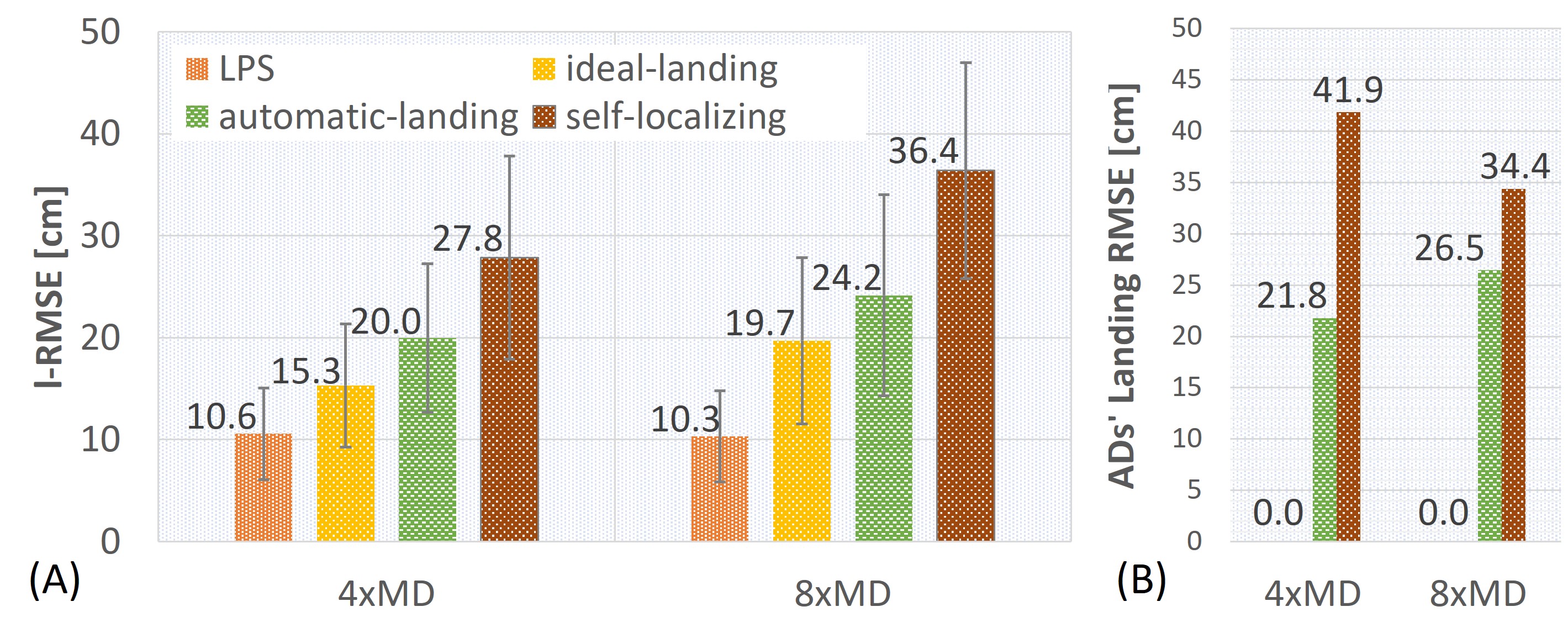}
    \caption{(A) Localization RMSE of the systems with 4 and 8 MDs. The mean error is reported. (B) Average AD landing error measured during experiments.}
    \label{fig:scalingMD_localization_performanc}
\end{figure}

\begin{figure}[b]
    \centering
    \includegraphics[width=1\columnwidth]
    {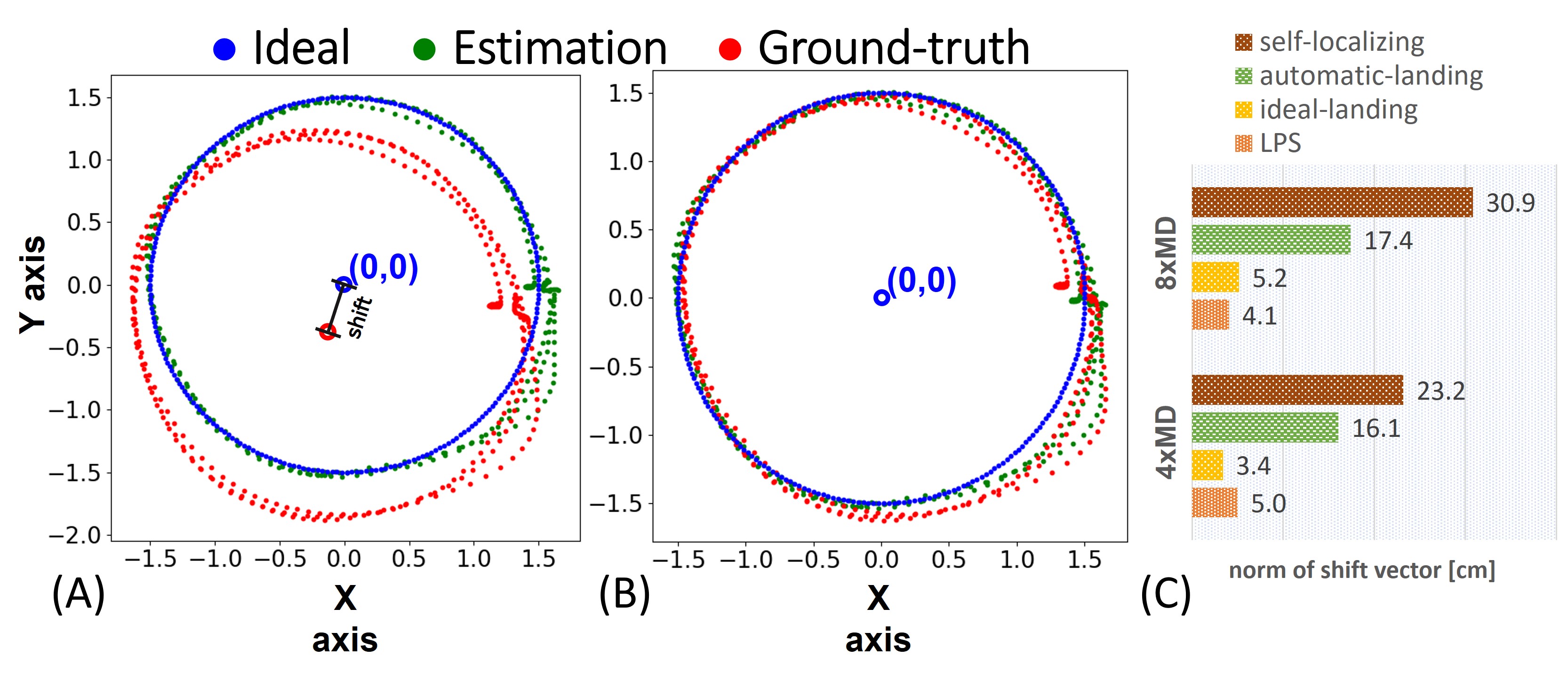}
    \caption{(A) MD trajectory as estimated on-board (green) and captured by the mocap (in red) with respect to the ideal (blue). (B) Comparison of trajectory shapes after alignment. (C) Norm of the shift vector.}
    \label{fig:shift_vector_and_circles}
\end{figure}

\subsection{Localization accuracy} \label{sec:localization_acc}

Our evaluation test-bed for the three proposed systems considers 4 ADs and 4 or 8 MDs. 
We evaluate the localization capacity during a 30-second flight of the MDs in a circular trajectory of a \SI{1.5}{\meter} radius with an average velocity of \SI{0.95}{\meter/\second}.
Drones fly at the height of \SI{0.8}{\meter}.
The trajectory is dictated by the available space of our indoor test arena and the coverage of the motion capture system.
We measure the average landing error $| p^A_{ADi} - \tilde{p}^T_{ADi} |$ of the ADs.
For every MD, we collected the onboard position estimate and the actual position, which is measured via mocap. 

Every experiment is repeated 5 times, and we report the mean and standard deviation values.  
We use two metrics to assess the MDs' localization capabilities, similar to previous evaluation frameworks for indoor localization systems~\cite{moron2022towards,zhao2021learning,Li2020RelativeLI}: a localization Root Mean Squared Error (l-RMSE) and a control RMSE (c-RMSE).
The l-RMSE accounts for the error between the position estimated by the MD (i.e., localization) and the actual one recorded by the mocap system.
The c-RMSE assesses how precisely the MDs can follow a desired trajectory, regardless of the global reference coordinates, e.g., how distant the trajectory is followed by every agent from an ideal circular trajectory. 
For comparison, we run the same experiments with the \textit{fixed-infrastructure} localization Commercial Off-The-Shelf (COTS) system, namely the \textit{Bitcraze Loco Positioning System}\footnote{https://www.bitcraze.io/documentation/system/positioning/loco-positioning-system/} (LPS), with  4 or 8 MDs and 6 ADs as recommended by their design specifications. 

In our experiments, we define the target landing positions of ADs as $p^T_{AD\mathrm{i}} = \{ (-2,2), (2,-2), (2,2), (-2,-2)  \} $.
Coordinates are expressed in meters according to a global frame.
ADs land on the ground.  
Within the automatic-landing and self-localizing scenarios, ADs take-off from $\{ (-1,0), (1,0), (0,1), (0,-1)\} $ before flying to the target points.
In the latter case, ADs run ASL just after the take-off.
For the LPS, we add two additional UWB anchors at coordinates $\{ (2, 0), (-2,0) \}$ and we position 3 of the UWB nodes at \SI{2.13}{\meter} height while the rest are placed on a \SI{0.17}{\meter} holder on the ground.

Fig.~\ref{fig:scalingMD_localization_performanc}-A reports the l-RMSE for all the considered systems with 4 and 8 MDs.
The ideal-landing system achieves a minimal l-RMSE of \SI{15.3}{\centi\meter} with 4 MDs.
This error slightly increases to \SI{19.7}{\centi\meter} with 8 MDs. 
We argue this effect to be caused 
by the sequential nature of TWR ranging that imposes extra time for the full fleet ranging as the number of MDs grows. Consequently, in a larger fleet, the UWB ranging information becomes less frequently available to the MDs and this contributes to higher localization error.
With 4 MDs, the automatic-landing and the self-localizing systems present a l-RMSE of \SI{20.0}{\centi\meter} and \SI{27.8}{\centi\meter}, respectively, which increase to \SI{24.2}{\centi\meter} and \SI{36.4}{\centi\meter} when using 8 MDs.
These designs present a lower localization performance with respect to the ideal-landing case because of the ADs' landing error. 
The displacement between the target and the real landing position achieves, on average, \SI{24.2}{\centi\meter} and \SI{38.2}{\centi\meter} for automatic-landing and self-localizing, respectively (see Fig.~\ref{fig:scalingMD_localization_performanc}-B).
The latter shows the highest displacement because the error estimate of the ASL routine propagates to the final landing position via the Kalman Filter. 
Compared with the LPS, which shows a l-RMSE of \SI{10.3}{\centi\meter} with 8 MDs, our systems present a higher l-RMSE up to 9.4, 13.9, and \SI{26.1}{\centi\meter} for the ideal-landing, automatic-landing, and self-localizing systems, respectively.
This corresponds to a localization error of 13.2\% in the self-localizing system as the most challenging system.

\begin{figure}[b]
    \centering
    \includegraphics[width=1\columnwidth]{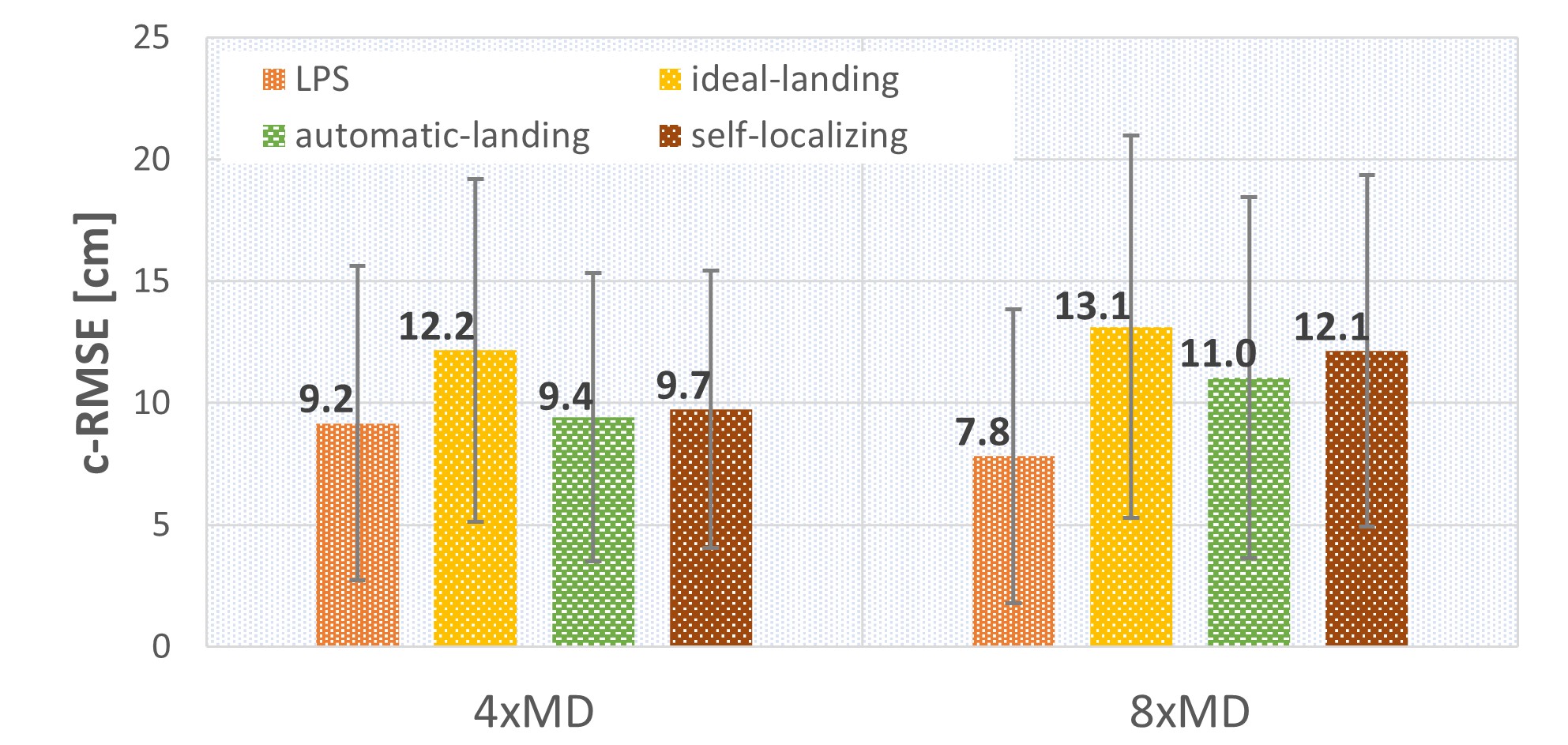}
    \caption{Control RMSE after eliminating the shift vector.}
    \label{fig:Translation_ideal_trajectory}
\end{figure}

\subsection{In-field control accuracy} \label{sec:control_acc}

To draw more insights into localization performance, the trajectories of individual MDs have been closely inspected. 
Fig.~\ref{fig:shift_vector_and_circles}-A
plots in green the on-board estimated locations of an MD within an automatic-landing experiment.  
In addition, we plot in blue the commanded circular trajectory and in red the real trajectory as measured by the mocap system. 
As can be observed from the figure, we see a translation effect between the estimated reference frame and real trajectories. 
However, this does not prevent the nano-drone from following the commanded circular trajectory, even when using the localization system with the highest l-RMSE.

We quantify the \textit{goodness} of the trajectories by computing the c-RMSE as the error between the shapes of the ideal trajectory (in blue) and the real trajectory (in red) after alignment, as shown in Fig.~\ref{fig:shift_vector_and_circles}-B.  
Fig.~\ref{fig:shift_vector_and_circles}-C plots the magnitude of the \textit{shift vectors} used to align the trajectories. 
These shifts are computed as the trimmed mean of the distance vectors between the estimated and real curves. 
We attribute the shifts to the origin, mainly from the AD's landing error, as their magnitude in case of automatic-landing and self-landing cases are up to 4.7$\times$ and 7.5$\times$ higher than the ideal landing case, where the shift is bounded to \SI{5}{\centi\meter}.
Concerning the c-RMSE reported in Fig.~\ref{fig:Translation_ideal_trajectory}, the studied designs show similar performance, suggesting that all the systems do not degrade the shape of the trajectories. 
In the case of 4 MDs, the c-RMSE amounts to 9.4-\SI{9.7}{\centi\meter} and slightly increases to 11-\SI{12.1}{\centi\meter} with 8 MDs. 
Therefore, neither scaling the number of MDs nor the large landing error prevent a swarm of nano-drones from following the commanded trajectory's shape.
%
When leveraging the \textit{self-localizing}, as the most flexible localization system, 
a nano-drone swarm can realize a flight formation by preserving distance between neighbor drones with a bounded relative distance error of 13.7\%, i.e., $\pm$\SI{15.6}{\centi\meter}.
To assess relative distance error, we compare the initial distance between drones and the distance maintained during the trajectory fly using mocap data.



\subsection{Onboard performance} \label{sec:onboard_perf}

The latency and energy costs of our localization systems refer to the UWB operations.
Every MD takes on average \SI{3}{\milli\second} for a DS-TWR exchange. 
Because of the fixed latency for DS-TWR, the ranging frequency is inversely proportional to the number of MDs. 
The average power increases with more MDs because a higher percentage of time is spent in RX mode, where we measure power of \SI{357}{\milli\watt}.
In our 4 and 8 MDs case studies, the MDs' UWB power overhead increases by 36.7\% and 42.8\%, respectively, compared to a single MD scenario, while a round of ranging is performed every \SI{48}{\milli\second} and \SI{96}{\milli\second}.
Concerning the 4 ADs, the UWB power cost during ranging accounts to \SI{150}{\milli\watt}. 
This cost represents 37\% of the total power(the motor power cost is zero after landing).

Concerning the self-localizing system, the initial ASL task takes \SI{5.7}{\second} in total. This latency accounts for the time to complete the ranging among all the anchors and the time to communicate the measurements to AD$0$.  
The execution of the MDS algorithm on the AD$0$ takes less than \SI{1}{\milli\second}.
More in detail, in our 4 AD use-case, AD$0$ completes the ranging with AD$1$-$3$ in \SI{890}{\milli\second} with a UWB power cost of \SI{224}{\milli\watt}. 
After ranging, AD$0$ takes an additional \SI{4.83}{\second} to aggregate results from the other ADs that send the measurements after ranging.
During the reception, the power of AD$0$ grows up to \SI{440}{\milli\watt} because of the RX power.
\section{Conclusion} \label{sec:conclusion}

This work addressed the limitation of static infrastructure setup for a UWB-based relative localization of multiple nano-drones.
We present three infrastructure-free design approaches by progressively relaxing the constraints on the known initial conditions of AD positions, which we also compare against a COTS baseline.
Our results, for a swarm of 4 ADs and 8 MDs, show a maximum localization error of up to \SI{27}{\centi\meter} higher than the COTS static ad-hoc infrastructure solution i.e., high flexibility of our system is gained with a reasonable cost in localization accuracy. 
Finally, our approach minimally impacts the control quality of the swarm while enabling dynamic and rapid deployment in any environment without relying on pre-existing infrastructure.

Future works may investigate the usage of the proposed
localization system in search and rescue tasks with bigger space to let the ADs fly further in the playground. 



\section{Acknowledgement} \label{sec:acknowledgement}
We thank the Center for Research on Complex Automated Systems and, in particular, Andrea Testa, and Lorenzo Pichierri for their support.


\bstctlcite{IEEEexample:BSTcontrol}

\bibliographystyle{IEEEtran}
\bibliography{IEEEabrv,bibliography}

\end{document}